\documentclass{article}

\usepackage{PRIMEarxiv}
\usepackage{multirow}
\usepackage[utf8]{inputenc} 
\usepackage[T1]{fontenc}    
\usepackage{hyperref}       
\usepackage{url}            
\usepackage{booktabs}       
\usepackage{amsfonts}       
\usepackage{nicefrac}       
\usepackage{microtype}      
\usepackage{lipsum}
\usepackage{fancyhdr}       
\usepackage{graphicx}       
\graphicspath{{media/}}     
\usepackage[most]{tcolorbox}
\title{A Hybrid Framework for Subject Analysis: Integrating Embedding-Based Regression Models with Large Language Models
\thanks{\textit{Citation:} This paper has been accepted at the 2025 ASIS\&T Annual Meeting. The final version will be published in the official proceedings.}

}

\author{
  Jinyu Liu \\
  University of North Texas \\
  Denton, TX, USA \\
  \texttt{JinyuLiu@my.unt.edu}
  \and
  Xiaoying Song \\
  University of North Texas \\
  Denton, TX, USA \\
  \texttt{XiaoyingSong@my.unt.edu}
  \and
  Diana Zhang \\
  University of North Texas \\
  Denton, TX, USA \\
  \texttt{DianaZhang@my.unt.edu}
  \And
  Jason Thomale \\
  University of North Texas \\
  Denton, TX, USA \\
  \texttt{Jason.Thomale@unt.edu}
  \And
  Daqing He \\
  University of Pittsburgh \\
  Pittsburgh, PA, USA \\
  \texttt{dah44@pitt.edu}
  \And
  Lingzi Hong \\
  University of North Texas \\
  Denton, TX, USA \\
  \texttt{Lingzi.Hong@unt.edu}
}

\begin{document}
\maketitle

\begin{abstract}
Providing subject access to information resources is an essential function of any library management system. Large language models (LLMs) have been widely used in classification and summarization tasks, but their capability to perform subject analysis is underexplored. Multi-label classification with traditional machine learning (ML) models has been used for subject analysis but struggles with unseen cases. LLMs offer an alternative but often over-generate and hallucinate. Therefore, we propose a hybrid framework that integrates embedding-based ML models with LLMs. This approach uses ML models to (1) predict the optimal number of LCSH labels to guide LLM predictions and (2) post-edit the predicted terms with actual LCSH terms to mitigate hallucinations. We experimented with LLMs and the hybrid framework to predict the subject terms of books using the Library of Congress Subject Headings (LCSH). Experiment results show that providing initial predictions to guide LLM generations and imposing post-edits result in more controlled and vocabulary-aligned outputs. 
\end{abstract}


\section{Introduction}

Subject analysis plays a crucial role in library information organization, serving as a foundational step in both cataloging and indexing processes \cite{salaba2023cataloging}. It involves examining the content of a resource to determine its core topics and assigning standardized subject terms\cite{sebastiani2002machine}. For example, a research article about the impact of climate change on agriculture may be assigned subject terms such as "climate change" and "agriculture”. The Library of Congress Subject Headings (LCSH) vocabulary is one such set of standardized subject terms widely used in libraries to help ensure consistency across their collections. The rate at which their collections tend to grow presents challenges for traditional manual subject analysis using LCSH \cite{ahmed2023automated}. Although machine and deep learning methods have been proposed to automate LCSH predictions\cite{kazi2021automatically}, these often struggle with issues such as data imbalance, class sparsity, high embedding cost, and limited generalization\cite{wehrmann2018hierarchical}\cite{tarekegn2021review}, which hinders the development of robust and scalable systems for automatic subject analysis. 

As more powerful AI models have emerged that are better suited for natural language processing – i.e., large language models (LLMs) – recent studies have begun employing them in subject analysis tasks
\cite{chow2024experiment}\cite{kluge2024few}. Although these studies have explored using different types of prompts, they have not investigated using more advanced LLM training techniques to tailor the models to the task at hand. For example, prompt engineering methods such as the Chain-of-Thought (CoT) have been shown to improve performance across several domain-specific tasks\cite{wei2022chain}. Fine-tuning LLMs using existing bibliographic datasets that already have LCSH terms assigned by library catalogers might also help improve their performance in domain-specific tasks like subject analysis\cite{rostam2024fine}. However, these methods have not yet been applied to subject analysis. 

Moreover, existing studies fail to address the limitations of LLMs. First, LLMs are limited in determining the appropriate number of subject terms, which often leads to over- or under-generation of terms. This is crucial, as it ensures a balanced representation of a resource’s topical scope, neither omitting key concepts nor overloading users with marginally relevant terms\cite{losee2004performance}. Second, LLMs often produce hallucinations, generating terms that do not come from the controlled vocabulary. To address these limitations, we propose to explore the following research questions:

(1) How can advanced LLM strategies, such as Chain-of-Thought prompting and fine-tuning, enhance performance in subject analysis tasks?

(2) How might the limitations of LLMs in subject analysis be mitigated?

To address these questions, we propose a hybrid framework incorporating more traditional machine learning (ML) models with LLMs to enhance the performance of LLMs in subject analysis. Specifically, to answer the first question, we adopt CoT reasoning and fine-tuning techniques and compare their performance against baseline models using direct zero-shot prompting. This allows us to assess the impact of these advanced LLM applications. To address the second question, we deploy ML models both before and after the LLM generation process. Leveraging their efficiency in numerical prediction, we use these models to estimate the optimal number of LCSH terms, which are then used to guide the LLM’s generation. Following the generation step, we apply a post-processing procedure to replace hallucinated terms with semantically similar entries from the LCSH vocabulary, enabling a potential improvement in accuracy. 

Our contributions are two-fold. First, we have systematically explored application and training strategies and evaluated how these techniques affect the performance of LLMs in subject terms generation. Second, we have proposed a novel hybrid framework that integrates embedding-based ML models with LLMs to guide subject term generation and post-processing. This framework not only improves LLMs’ generation accuracy and efficiency but also ensures more reliable and standard outputs. Our framework offers insights toward a practical and scalable solution for using AI to assist the subject analysis task.

\section{RELATED WORK}
\subsection{Application of Traditional Machine Learning and Deep Learning in Subject Analysis}

AI techniques have been widely used in bibliographic metadata processing \cite{okunlaya2022artificial}\cite{cox2024defining}. Yu et al.(\cite{yu2022survey} focused on knowledge-enhanced generation using ML and deep learning (DL) models. ML models like support vector machines (SVM) and tree-based architecture (random forest, decision tree) have been used in text extraction and subject assignment\cite{sharma2023machine}. While these ML models perform well on structured data, they cannot handle high-dimensional classification like LCSH prediction. 

DL models like neural networks have been used in text prediction and semantic understanding of document sets. Kowsari et al.\cite{kowsari2017hdltex} applied DL architectures for multi-level classification. They demonstrated that these DL models outperform conventional ML classifiers such as Bayes and SVM. Kandimalla et al.\cite{kandimalla2021large} proposed a Deep Attentive Neural Network (DANN) for subject classification. They trained DANN on 9 million samples, and their model achieved an F1 score of 0.76. Although these models can capture deeper features and relationships within a dataset, they need a large training dataset to ensure good performance. Some recent studies have applied pre-trained DL models to library metadata. Bidirectional Encoder Representations from Transformers (BERT) and domain-specific versions (SciBERT) have demonstrated powerful capabilities in bibliographic metadata classification \cite{zhang2023survey}. George and Sumathy \cite{george2023integrated} offered a clustering BERT-LDA framework to improve the topic extraction model. Although DL models that have been pre-trained on authoritative and relevant datasets perform well and improve upon conventional ML techniques, they still exhibit issues like data imbalance, class sparsity, and the high cost of updating their embeddings.

Because a knowledge object often belongs to multiple subject categories, subject analysis can be considered a multi-label classification (MLC) problem, which ML and DL methods have been explored to solve. Han et al.\cite{han2023survey}, divided MLC tasks between supervised and semi-supervised learning methods. They evaluated various methodologies, including decision trees, Bayes, SVMs, neural networks, and KNN-based models. Extreme multi-label classification (EMLC) can also be implemented in automatic cataloging systems\cite{rajendram2024advancing}. Like pre-trained models, the models designed for MLC problems have difficulty in handling class sparsity and high-dimensional label space. Many labels are used at low frequency (appearing once in the whole dataset), which leads to an imbalance in the training data. Additionally, the large number of labels or categories makes the classification cost computationally expensive.

\subsection{LLMs’ Application and Disadvantages}
LLMs offer strong contextual understanding and powerful versatility. Adetayo \cite{adetayo2023artificial} explored the implementation of LLMs in academic libraries. Due to their strong transfer learning abilities, LLMs perform well when trained with minimal or even no finetuning \cite{lin2024data}. However, LLMs still present many limitations. First, they are trained on general datasets, so they may fail to perform well in specialized tasks like subject analysis.  Chow et al. \cite{chow2024experiment} showed that GPT-3.5 can generate mostly correct (90\%) MARC coding but only 23.3\% valid LCSH for ETD records. Second, most existing applications of LLMs in cataloging rely on prompt engineering and fail to explore training or adaptation techniques \cite{chataut2024comparative}\cite{maragheh2023llm}\cite{zhang2023utilising}. Finally, while training strategies such as fine-tuning can help improve task alignment, they do not fully solve hallucination issues in LLMs \cite{dunn2022structured}. Because invalid or incorrectly applied subject headings are useless, LLMs cannot perform subject analysis well if hallucinations and lack of domain applicability cannot be addressed.

\section{METHODOLOGY}
\subsection{Overview}
 Although traditional ML models and LLMs each have weaknesses that make them each suboptimal on their own for subject analysis, they do have complementary strengths. On the one hand, LLMs often lack precise control over output structure and quantity, and they are prone to hallucinations, but they excel at generating natural language. On the other hand, traditional ML models are less flexible than LLMs, but they offer greater interpretability, data efficiency, and controllability. They are suited for predicting structured outputs, such as estimating the number of subject headings based on features extracted from metadata (e.g., titles and abstracts). We hypothesize that these predictive numbers can be used to guide LLM generation, providing constraints that help prevent over-generation or under-generation. We propose a hybrid framework that combines the strengths of traditional, lightweight predictive models with those of LLMs, while adopting multiple strategies to improve the performance of LLMs in automatic subject analysis (Figure 1). The small ML models serve as a guide and constraint for the LLMs. Meanwhile, LLMs afford powerful generalization and contextual reasoning.
 
The framework includes three main phases. In the first phase, the baseline performance of LLMs in subject analysis is evaluated through zero-shot learning, and CoT reasoning and fine-tuning are explored to further enhance their reasoning ability. In the second, to address the issue of label quantity inconsistency, we apply small-scale predictive models trained on structured metadata (title and abstracts) to estimate the optimal number of LCSH terms, and the predicted number is used to guide what the LLM generates. In the third, we implement a post-processing step for replacing generated terms not in the LCSH vocabulary with their most semantically similar LCSH equivalent.

\begin{figure}[htbp]
    \centering
    \includegraphics[width=0.9\linewidth]{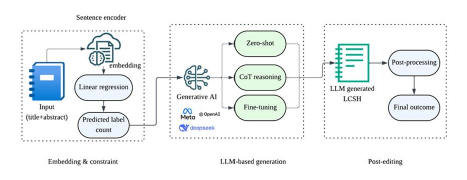}
    \caption{Method Overview}
    
\end{figure}

\subsection{Phase One: Apply LLMs Only to Generate Library of Congress Subject Headings}
\subsubsection{Prompt Engineering Methods}

(1) Prompt Engineering Methods
We first conducted a set of experiments using zero-shot and few-shot learning to evaluate the baseline performance of LLMs in subject terms assignments. In our framework, a task-specific prompt was designed to instruct the LLM to generate relevant LCSH subject headings based on the input title and abstract. The prompt included a role instruction defining the “assistant” persona and their general objective, a clear task definition, and a description of the expected output format. In few-shot learning, we provided examples of subject analysis samples. Each example included the title, abstract, and the LCSH subject headings assigned by library catalogers. 
Here is an example of the zero-shot learning prompt:

\begin{tcolorbox}[colback=blue!5!white, colframe=blue!60!black, boxrule=0.8pt, sharp corners=southwest, fonttitle=\bfseries]
"You are a helpful assistant predicting Library of Congress Subject Headings (LCSH) for books."

"Given the title and abstract, predict relevant LCSH labels."

"Respond only with the predicted LCSH labels separated by commas."
\end{tcolorbox}
Conventional LLM-based generation typically contains only a single forward generation pass. This means all outputs are generated at once without an iterative process. While computationally efficient, this often causes redundancy or insufficient outputs. To mitigate these issues, we applied a CoT reasoning approach. By designing multiple prompts, we guided the model to infer the LCSH terms through a step-by-step reasoning process rather than generating all labels in one pass. This multi-round generation enables models to consider previously generated labels when predicting subsequent ones, which may enhance term diversity and conceptual coverage, making results more accurate and systematic. To demonstrate how we adapted CoT reasoning adapted for subject analysis, consider a book titled “Museums in China: Materialized Power and Objectified Identities” and its abstract describing the relevant environmental and background information. We would guide the LLM to generate labels iteratively, prompting it to reason about different aspects of the item through multiple rounds. In the first round, the LLM might focus on the environmental and anthropological context, predicting subject terms like “Anthropology, China.” In the second round, the LLM considers both the original input and the previously generated terms to predict new terms focusing on different aspects, like economic and social. The LLM might generate “Colonial influence, Heritage tourism, Cultural property, Ethnology.” In the last round, we ask the LLM to generate additional new terms, considering the results of the previous two rounds. The LLM might now give “Identity (Psychology), Museum curators, Museums, Nationalism, Power (Social sciences), Social change” as a result. All generated terms from the three rounds are then combined into a candidate list of LCSH subject headings for the book. One key advantage of CoT is its controllability. The number of rounds and labels generated at each stage can be easily adjusted. This allows the CoT structure to meet specific task requirements. For example, users can define the number of iterations needed to meet the comprehensiveness of targets or set constraints on the output size per round.

Here is an example of the CoT prompt:

\begin{tcolorbox}[
  colback=blue!3!white,
  colframe=blue!80!black,
  boxrule=0.8pt,
  sharp corners,
  enhanced,
  fontupper=\small\ttfamily, 
]
"You are a helpful assistant predicting Library of Congress Subject Headings (LCSH) for books."

\textcolor{gray}{\#First round inference}

"Predict exactly \{n\} Library of Congress Subject Headings (LCSH) labels."

\textcolor{gray}{\#Second round inference}

"Predict \{2*n\} additional Library of Congress Subject Headings (LCSH) labels that explore different aspects. Avoid repeating or closely related labels compared to the current ones."

\textcolor{gray}{\#Third round inference}

"Predict additional Library of Congress Subject Headings (LCSH) labels. Avoid repeating or closely related labels compared to the current ones."

"Respond only with the predicted LCSH labels separated by commas."
\end{tcolorbox}

\subsubsection{Finetuning Methods}
Fine-tuning is a powerful technique that enables LLMs to learn and adapt to domain-specific professional knowledge, significantly enhancing their ability to generate accurate and contextually appropriate outputs\cite{zhang2024gpt4roi}. Our experiments used structured metadata containing standard LCSH terms assigned by library catalogers to fine-tune our models and address the limitations of zero-shot learning. Two complementary fine-tuning strategies were explored: Supervised Fine-Tuning (SFT)\cite{ouyang2022training} with full parameters and Low-Rank Adaptation (LoRA) \cite{hu2022lora}. The SFT, full parameter setting updated all LLM weights. In contrast, LoRA froze the original weights, implementing layers of trainable, low-rank matrices to reduce the memory and training time burden. SFT allows the model to learn mappings directly between the structured inputs (titles, abstracts) and the expected LCSH outputs, whereas LoRA allows fine-tuning using significantly fewer parameters and GPU resources. 

The models were trained with the metadata input (abstract and title) and the ground truth LCSH subject terms. We hoped such training would allow the models to better understand the idiosyncratic structure and vocabulary of the subject analysis task improving the domain-applicability of its output and reducing hallucination issues.

\subsection{Phase Two: Enhance Prompt Engineering with the Optimal Number of LCSH labels}
To encourage an LLM to produce an appropriate number of subject terms, we hypothesized that external guidance from smaller, task-specific models could be beneficial. We explored the use of lightweight regression and embedding models to predict the optimal number of LCSH terms. In subject analysis, the number of relevant subject headings varies across samples. To estimate the optimal number for a given sample, embedding models first convert the input text to vector representations to capture the deeper contextual meaning. Based on these embeddings, ML models are built to predict how many LCSH terms should be assigned to a sample. The predicted number is then used to guide and constrain LLMs' outputs. 

\subsection{Phase Three: Address LLM Hallucinations with Post-processing}
A post-processing step was adopted to attempt to further mitigate hallucinations. Specifically, hallucinated terms that appear plausible but are not present in the LCSH-controlled vocabulary were identified and replaced with their most semantically similar ones from LCSH. This correction mechanism ensures that all generated subject headings belong to the set of standardized LCSH terms. Using a post-processing approach enhances both the usability and accuracy of the framework without making any modifications to the LLM architecture itself.

\section{EXPERIMENT DETAILS}
\subsection{Data Preparation}
Our dataset was provided by the University of North Texas (UNT)\cite{untlibrary} Library Catalog which gave us 100,000 samples in Machine-Readable Cataloging (MARC21) format. To prepare the data, we first converted the raw MARC to JSON format using a Python script. Then we extracted the desired fields: the title (MARC 245), summary or abstract (MARC 520), Library of Congress Classification (LCC) (MARC 050), and Library of Congress Subject Headings (LCSH) (MARC 650, subfield “\$a” only). After extraction, preprocessing steps were applied to clean the data, including removing duplicates, converting text to lowercase for normalization, handling missing values, and processing special characters. Following the cleaning process, we retained 78,260 valid samples.

After preprocessing the dataset, we then split it into a training and testing set. For the testing set, we selected 100 samples from each of the 21 LCC categories. This was used for benchmarking experiments, including zero-shot and CoT evaluations. The remaining 76,160 samples compose the training set, used to fine-tune LLMs. 

\subsection{Zero-Shot/Few-shot Learning }
To evaluate the baseline performance of different LLMs for subject analysis, we used zero- and few-shot learning prompts to generate LCSH labels. This allowed us to examine the inherent capability of each model and assess the suitability of different base architectures to the task. The LLMs  evaluated include Llama-3.1-8b, GPT-3.5, GPT-4, DeepSeek-V3 and DeepSeek-R1. Llama-3.1-8b is a variant of Llama 3\cite{grattafiori2024llama}  provided by Meta AI. It only has 8 billion parameters. GPT-3.5 is an intermediate model developed by OpenAI. GPT-3.5 serves as a bridge between GPT-3 and GPT-4. GPT-4 \cite{achiam2023gpt} is a state-of-the-art large language model offered by OpenAI beginning in 2023. DeepSeek V3 \cite{liu2024deepseek} is a large chat language model that was provided by DeepSeek AI in 2024. Their benchmark tests demonstrated that DeepSeek V3 outperforms models like Llama 3.1 and Qwen 2.5 \cite{yang2025qwen3}. DeepSeek R1\cite{guo2025deepseek} is an advanced AI reasoning model developed by DeepSeek and published in 2025. Although both DeepSeek V3 (671 billion parameters) and DeepSeek R1 (685 billion parameters) are open-source models, their massive scale requires significant hardware and computational costs. This makes it extremely challenging to deploy and run efficiently on our machines. Unlike the local deployment of Llama-3.1-8b, the GPT-based models  and DeepSeek-based models  are accessed via their respective APIs. 

\subsection{Chain of Thought (CoT)}

To evaluate the effectiveness of CoT in enhancing LCSH label generation, we designed a three-round multi-turn inference strategy using LLaMA-3.1-8B. The experiments ran on a local server using the Hugging Face Transformers and PEFT libraries in FP16 precision mode. Each test sample was composed of a title and abstract. In Round 1, the LLM was prompted to generate an initial set of subject headings based on the title and abstract alone. In Round 2, the prompt was updated to include both the original metadata and the terms generated in Round 1. In Round 3, the LLM was asked to generate as many additional relevant terms as possible, aiming for comprehensive coverage. All outputs from the three rounds were then aggregated and used as the final prediction set.

To explore how different prompting strategies influence the model's performance, we experimented with three prompt variants (V1, V2, and V3) in the CoT setting. V2, the most complex and detailed version, explicitly instructed the LLM to generate LCSH terms that differ from those already produced and to focus on unexplored aspects. V3, a moderately simplified version, encouraged the generation of terms from distinct directions but did not avoid non-repetition.  

\subsection{Fine-tuning }

For the fine-tuning stage, we adopted the same prompt structure used in zero-shot learning to ensure consistency. The whole training set (76,160 samples) was used for this purpose. Fine-tuning was implemented using the Supervised Fine-Tuning (SFT) framework provided by Hugging Face. A series of experiments were conducted by adjusting key hyperparameters, including the number of epochs (1, 3, 5, 10), learning rate (1e-4, 5e-5, 1e-5), and temperature (0.7, 1.0) to examine their impact on model performance. Each configuration was evaluated on the same 2,100 sample test set using precision, recall, and F1-score. This allowed us to analyze how different configurations affect evaluation metrics and output quality. Considering constraints like model openness, deployment scalability, and computational costs, we utilized only the Llama 3.1-8B model  in CoT and fine-tuning.

\subsection{Predicting the Number of LCSH Terms for Constrained Label Generation}
The appropriate number of LCSH terms differs depending on the content they describe. To accommodate different samples and generate accurate results, we built ML models to predict the optimal number of labels for each sample . First, the input data (title and abstract) was encoded as numerical representations using pre-trained language models. We adopted BERT \cite{devlin2019bert} and SciBERT \cite{beltagy2019scibert} as the embedding models. These transformer-based models\cite{han2021transformer}  have powerful contextual understanding abilities\cite{amatriain2023transformer}\cite{kalyan2021ammus}. The embeddings were then used as input features for downstream ML models trained to predict the expected number of subject terms for a sample. The ML models we used include Random Forest (RF)\cite{breiman2001random}, Linear Regression \cite{james2023linear}, and XGBoost \cite{chen2016xgboost}. These models were trained on the training set and applied to the testing set. The best-performing model was then used to predict the appropriate number of subject headings to guide further LLM generation experiments. Specifically, for samples in the testing set, the predicted number N was incorporated into the prompt design and served as a constraint.

To explore the trade-off between recall and precision, we further conducted experiments using multiple constraint values (N, 2N, 3N) to guide LLM output. While N provides an estimate of the optimal number of LCSH labels, expanding this value in controlled experiments allowed us to explore how increasing the number of candidate labels affects LLM performance. 

\subsection{Post-processing to Substitute Hallucinations with Valid LCSH Terms}
We experimented with the post-processing step to enhance label validity and reduce hallucinations. This step identified terms generated by the LLM that do not exist in the LCSH controlled vocabulary and then replaced them with the most semantically similar terms. First, we obtained the LCSH file from the Library of Congress website. All subject headings from MARC field 650 \$a were extracted, cleaned, and deduplicated to ensure that unique and standardized labels were retained. After preprocessing, the vocabulary contained approximately 318,500 unique LCSH terms. To enable efficient semantic matching, we used the all-mpnet-base-v2  pre-trained BERT-based embedding model to encode all controlled vocabulary terms into high-dimensional vectors. However, the resulting embeddings were 628-dimensional, which is computationally challenging for large-scale similarity search. To handle this, we implemented Principal Component Analysis (PCA) \cite{hotelling1933analysis}, which can reduce the number of features and still maintain important information in the data, reducing the computational cost. In our case, PCA reduced the embedding dimensionality from 628 to 50. Then we applied Nearest Neighbor Search (NNS)\cite{andoni2018approximate} to identify the most similar LCSH term for each non-standard label produced by the LLM. NSS was implemented by calling Facebook AI Similarity Search (FAISS)\cite{douze2024faiss} . Figure 2 shows the post-processing workflow. 

\begin{figure}[htbp]
    \centering
    \includegraphics[width=0.9\linewidth]{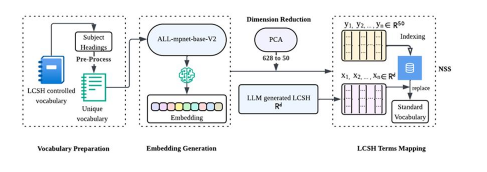}
    \caption{The Workflow of Post-processing}
    
\end{figure}

\section{EXPERIMENT RESULTS AND ANALYSIS}

To evaluate LCSH term generation, we primarily use recall, precision, F1-score, and the average number of generations as key metrics. These metrics help assess the accuracy of predictions and the trade-off between label counts and output redundancy.

\subsection{Evaluation of LLM-Only Methods in Subject Analysis}
\subsubsection{Prompt Engineering Methods}

We initially experimented with both zero-shot and few-shot prompts to determine which would perform better as a baseline. Few-shot prompts performed worse, probably due to lack of information to guide the output, so we applied the zero-shot prompt with the same generation parameters across all LLMs to compare their baseline performance. Despite using the same settings, we observed significant differences in the output between models (Table 1). The Llama-based and DeepSeek-based models produce a similar number of predictions on average, from 13 to 15 terms per sample. In contrast, the GPT-based models generate fewer labels, on average, from 5 to 8 per sample. Among these models, DeepSeek-R1 achieved the highest recall (69\%). GPT-3.5 achieved the highest precision and F1-score due to its lower number of generated outputs. These results demonstrate that LLM architecture does impact model performance. However, due to budgetary constraints, we had to use Llama-3.1-8b for the rest of our experiments. (These results could provide insights for applying other LLMs.) 

\begin{table}[ht]
\centering
\caption{Various LLMs’ Performance in Zero-shot Learning}
\begin{tabular}{lcccc}
\toprule
\textbf{LLMs} & \textbf{Recall} & \textbf{Precision} & \textbf{F1} & \textbf{Average Number of Terms} \\
\midrule
Llama-3.1-8b  & 0.43 & 0.08 & 0.135 & 14.89 \\
GPT-4         & 0.59 & 0.20 & 0.299 & 7.58 \\
GPT-3.5       & 0.47 & \textbf{0.24} & \textbf{0.318} & 5.56 \\
DeepSeek-V3   & 0.32 & 0.06 & 0.101 & 13.39 \\
DeepSeek-R1   & \textbf{0.69} & 0.14 & 0.233 & 13.38 \\
\bottomrule
\end{tabular}
\label{tab:llm-zero-shot}
\end{table}

The Chain-of-Thought (CoT) approach using several rounds of predictions significantly enhanced the diversity and richness of LCSH generation. For example, the book titled “Museum Rhetoric: Building Civic Identity in National Spaces” has subject headings of “museums; national museums; rhetoric; nationalism; cultural property.” Llama-3.1-8b with zero-shot learning generated the terms, “Museums, Rhetoric, Cultural studies, Identity formation, Social aspects of Museums, Cultural policy, Civic Engagement, National identity, Cultural heritage, Public spaces” and achieved 0.2 recall. In contrast, Llama-3.1-8b with CoT predicted the terms, “Civic engagement, Cultural heritage, Cultural policy, Heritage conservation, Identity (Psychology), Museums, Nationalism, Patriotism, Place (Philosophy), Public spaces, Rhetoric, Symbolic expression, Urban planning and culture” and achieved 0.4 recall. The CoT-generated LCSH labels cover more aspects of the input content. Notably, how the prompts are formulated appears to affect performance (Table 2). Despite having more detailed instructions in V2 and V3, the V1 design with simpler instructions produced better results. 

\begin{table}[ht]
\centering
\caption{Results of CoT Prompt Versions}
\begin{tabular}{llcccc}
\toprule
\textbf{LLM} & \textbf{Prompt} & \textbf{Recall} & \textbf{Precision} & \textbf{F1} & \textbf{Average Number of Terms} \\
\midrule
\multirow{3}{*}{Llama-3.1-8b} 
  & V1 & \textbf{0.51} & 0.04 & 0.074 & 32.83 \\
  & V2 & 0.43 & 0.04 & 0.073 & 29.52 \\
  & V3 & 0.46 & 0.03 & 0.056 & 34.05 \\
\bottomrule
\end{tabular}
\label{tab:cot-prompts}
\end{table}

\subsubsection{Fine-tuning}
Table 3 shows the results of fine-tuning. LoRA-based fine-tuning significantly outperformed full-parameter finetuning (SFT) in terms of both efficiency and usability. In SFT, all weights in the LLaMA-3.1-8B model must be updated (nearly 55 GB) and the entire model must be saved after fine-tuning. In contrast, LoRA fine-tuning requires updating and storing only a small set of parameters (nearly 155 MB), while we still need the base Llama-3.1-8b model during inference. In terms of training time, SFT needs approximately 12 hours to complete a single epoch, while LoRA reduces this time to 3 hours for one epoch.

The more epochs used to train a model, the better it performed. Results show  that recall increased from 33\% after 1 epoch of training to 52\% after 10 epochs of training. More epochs enable the model to better understand domain-specific tasks and enhance its ability to capture profound relationships between inputs and outputs. However, the training time and computational costs also increase. LoRA’s improved efficiency allowed us to train many more epochs quickly with fewer resources compared to SFT, producing better results.

\begin{table}[ht]
\centering
\caption{Fine-tuning with Llama-3.1-8b}
\begin{tabular}{llcccc}
\toprule
\textbf{LLM} & \textbf{Parameter} & \textbf{Recall} & \textbf{Precision} & \textbf{F1} & \textbf{Average Number of Terms} \\
\midrule
\multirow{8}{*}{Llama-3.1-8b}
  & SFT, epoch=1, token=150      & 0.13 & 0.14 & 0.135 & 1.94 \\
  & SFT, epoch=3, token=150      & 0.29 & 0.04 & 0.070 & 21.56 \\
  & LoRA, epoch=1, token=150     & 0.33 & 0.05 & 0.087 & 23.21 \\
  & LoRA, epoch=3, token=150     & 0.41 & 0.06 & 0.105 & 24.46 \\
  & LoRA, epoch=5, token=150     & 0.49 & 0.04 & 0.074 & 38.37 \\
  & LoRA, epoch=10, token=150    & \textbf{0.52} & 0.06 & 0.108 & 33.76 \\
  & LoRA, epoch=10, token=100    & \textbf{0.52} & 0.07 & 0.123 & 24.73 \\
  & LoRA, epoch=10, token=50     & 0.48 & 0.09 & \textbf{0.152} & 14.11 \\
\bottomrule
\end{tabular}
\label{tab:fine-tuning-llama}
\end{table}

We also controlled the number of generated LCSH labels using the ‘max-new-tokens’ (“token”) parameter. This parameter defines the maximum number of tokens the LLM is allowed to generate during inference. By adjusting this, we can effectively manage the balance between recall and precision. A higher “max-new-tokens” value enables the model to produce a more comprehensive set of LCSH terms to achieve a higher recall. For example, when the parameter is set to 150, the model generates an average of 33.76 terms per sample and achieves a high recall of 52\%. However, this increase in recall often leads to lower precision. Our experiment demonstrates that adjusting “max-new-tokens” is a simple but effective strategy for balancing recall and precision in fine-tuned LLMs. 

\subsection{Guiding LLMs with Optimal Number of LCSH Terms Constraint}
\subsubsection{Predict the number of LCSH Terms}

When attempting to predict the appropriate number of LCSH terms for a sample, we found that the BERT-based embedding model (All-mpnet-base-v2) outperformed SciBERT (Table 4). Among the various model combinations, the BERT-based model with linear regression achieved the lowest average difference between predicted and actual label counts. The combination of the BERT-based model with XGBoost showed the highest Pearson Correlation Coefficient (PCC) and the lowest Root Mean Square Error (RMSE) values. We selected the combination of linear regression with All-mpnet-base-v2 embedding models for the prediction of the optimal number of LCSH terms.

\begin{table}[ht]
\centering
\caption{Results of Prediction of the Number of LCSH Labels}
\begin{tabular}{lllccc}
\toprule
\textbf{Input} & \textbf{Embedding model} & \textbf{Prediction model} & \textbf{Average Difference} & \textbf{PCC} & \textbf{RMSE} \\
\midrule
\multirow{3}{*}{\textbf{Title + Abstract}} 
  & \multirow{3}{*}{All-mpnet-base-v2}
    & Linear regression & \textbf{1.093} & 0.370 & 1.528 \\
  & & RF                & 1.148 & 0.272 & 1.580 \\
  & & XGBoost           & 1.096 & \textbf{0.378} & \textbf{1.517} \\
\cmidrule{2-6}
  & \multirow{3}{*}{SciBERT}
    & Linear regression & 1.106 & 0.348 & 1.541 \\
  & & RF                & 1.150 & 0.283 & 1.575 \\
  & & XGBoost           & 1.099 & 0.371 & 1.523 \\
\bottomrule
\end{tabular}
\label{tab:lcsh-prediction}
\end{table}

\subsubsection{Constrain the Number of LLM Outputs}

Table 5 presents the performance of the LLaMA-3.1-8B model under different output constraints. The zero-shot baseline didn’t apply any constraint on the number of generated labels. It achieved the highest recall (43\%) but suffered from low precision (8\%) and F1-score (13\%). On average, it generated 14.89 LCSH labels per sample and often included many irrelevant terms. This means that the model without constraint tends to over-generate, often including noisy or irrelevant LCSH terms to ensure coverage. Once we included N to guide LLM generation, the average number of output terms dropped significantly to 3.14, which is a 79\% reduction in output size. Despite this reduction, precision improved from 0.08 to 0.21, and the F1 score reached its highest value (0.24). This indicates that constraining generation with a predictive number of LCSH terms improves the relevance and focus of predictions while still preserving meaningful recall (0.29). As we relax the constraint (Limit 2N, Limit 3N): output size increases gradually from 5.58 to 8.26, with recall rising again from 0.37 to 0.41. However, precision declines from 0.16 to 0.12, and F1 is lower than Limit N. 

The Average Number of Terms directly reflects the model's generalization and its risk of hallucination. Unconstrained output (14.89) almost leads to irrelevant or redundant terms. Constrained settings (3.14, 5.58, 8.26) are more aligned with real-world subject analysis use cases (typically 1–6 per record in libraries). These results demonstrate that controlling the number of generated outputs provides an effective approach to balance recall and precision. In scenarios where comprehensive coverage of subject terms is essential like academic indexing or exploratory search, higher recall is preferred, even at the expense of precision. In such cases, allowing the model to generate a larger set of candidate labels (e.g., using 2N or 3N as a constraint) ensures that more relevant terms are captured. Conversely, when the goal is highly precise pursuit such as for professional catalog records or metadata-driven recommendation systems. Limiting the number of generated terms (e.g., using N) helps reduce noise.

\begin{table}[ht]
\centering
\caption{Constrain the number of labels an LLM generates}
\begin{tabular}{llcccc}
\toprule
\textbf{LLMs} & \textbf{Methods} & \textbf{Recall} & \textbf{Precision} & \textbf{F1} & \textbf{Avg. \# Terms} \\
\midrule
\multirow{4}{*}{Llama-3.1-8b}
  & Zero-shot   & \textbf{0.43} & 0.08 & 0.135 & 14.89 \\
  & Limit n     & 0.29 & \textbf{0.21} & \textbf{0.244} & 3.14 \\
  & Limit 2n    & 0.37 & 0.16 & 0.223 & 5.58 \\
  & Limit 3n    & 0.41 & 0.12 & 0.186 & 8.26 \\
\bottomrule
\end{tabular}
\label{tab:limit-labels}
\end{table}

Table 6 presents the results using the optimal number of subject terms (n) to guide CoT inference. We observed that using “n” to guide the model generated fewer terms, giving a relatively high recall (0.45) and precision (0.10). Asking the LLM to generate as many as possible (AMAP) slightly increases recall from 0.45 to 0.51, but there are too many irrelevant terms in the final set. Finally, increasing the number of reasoning rounds beyond three does not lead to further improvements in recall.

\begin{table}[ht]
\centering
\caption{Apply Constraints to CoT Prompts}
\small 
\begin{tabular}{llllcccc}
\toprule
\textbf{LLMs} & \textbf{Methods} & \textbf{Rounds} & \textbf{\# Subject Terms Each Round} & \textbf{Recall} & \textbf{Precision} & \textbf{F1} & \textbf{Avg \# Terms} \\
\midrule
\multirow{4}{*}{Llama-3.1-8b}
  & \multirow{4}{*}{CoT}
  & 3 & 2/3/5           & 0.43 & 0.09 & 0.149 & 10.94 \\
  & & 3 & n/n/2n        & 0.45 & 0.10 & \textbf{0.164} & 11.10 \\
  & & 3 & n/2n/AMAP     & \textbf{0.51} & 0.04 & 0.074 & 32.83 \\
  & & 4 & n/n/2n/AMAP   & \textbf{0.51} & 0.04 & 0.074 & 36.07 \\
\bottomrule
\end{tabular}
\label{tab:cot-constraint}
\end{table}

\subsection{Vocabulary Mapping with LCSH Terms}
Table 7 demonstrates the effectiveness of our proposed strategies in improving LLMs’ prediction performance in subject analysis. By applying fine-tuning and CoT, we can achieve an 8\% (0.43->0.51) and 9\% (0.43->0.52) increase in recall before applying the post-processing techniques. However, as the number of the generated outputs increases, both precision and F1-score are reduced. This confirms that constraining the number of generated labels effectively balances precision and recall and mitigates the risk of over-generation.

After applying the vocabulary mapping with LCSH terms, we further achieved a 12\% (0.51->0.63) and 9\% (0.43->0.52) improvement in recall for CoT and zero-shot. This 9\% to 12\% improvement in recall highlights the effectiveness of the vocabulary mapping. The precision also increases as the hallucinated terms are replaced with the correct LCSH terms. The vocabulary mapping method leads to better recall and precision at the same time, proving its value in automated subject analysis. 

In summary, the combined use of small ML models and LLMs using advanced prompt-engineering techniques leads to significantly better results: the recall increased from 43\% (zero-shot baseline) to 63\% (CoT with post-processing), and the precision improved from 8\% (zero-shot baseline) to 0.26 (limit n with post-processing). 

\begin{table}[ht]
\centering
\caption{Result in Llama-3.1-8b Before and After Post-processing}
\begin{scriptsize}
\begin{tabular}{llcccccccc}
\toprule
\textbf{LLM} & \textbf{Methods} 
& \textbf{Recall (B)} & \textbf{Recall (A)} 
& \textbf{Precision (B)} & \textbf{Precision (A)} 
& \textbf{F1 (B)} & \textbf{F1 (A)} 
& \textbf{Avg Terms (B)} & \textbf{Avg Terms (A)} \\
\midrule
\multirow{6}{*}{Llama-3.1-8b}
  & Zero-shot   & 0.43 & 0.52 & 0.08 & 0.09 & 0.135 & 0.155 & 14.89 & 14.69 \\
  & CoT         & 0.51 & \textbf{0.63} & 0.04 & 0.05 & 0.074 & 0.091 & 32.83 & 30.26 \\
  & Fine-tuning & \textbf{0.52} & 0.57 & 0.06 & 0.06 & 0.108 & 0.111 & 33.76 & 33.11 \\
  & Limit n     & 0.29 & 0.36 & \textbf{0.21} & \textbf{0.26} & \textbf{0.244} & \textbf{0.300} & 3.14 & 3.23 \\
  & Limit 2n    & 0.37 & 0.44 & 0.16 & 0.19 & 0.223 & 0.265 & 5.58 & 5.67 \\
  & Limit 3n    & 0.41 & 0.48 & 0.12 & 0.14 & 0.186 & 0.217 & 8.26 & 8.29 \\
\bottomrule
\end{tabular}
\label{tab:postprocessing}
\end{scriptsize}
\end{table}

Notably, the effectiveness of post-processing varied between methods. For fine-tuned models, recall improved by only approximately 5\% and precision did not improve at all after post-processing, whereas zero-shot and CoT approaches showed improvements in recall of 9\% and 12\%, respectively, and a 1\% improvement in precision. This may reflect that fine-tuned models have learned to generate more standard LCSH terms during training. Whereas, relying only on prompt engineering tends to produce terms that are more diverse but hallucinated. Untrained LLMs may benefit more from vocabulary mapping during post-processing. 

\section{LIMITATIONS}
Our study exhibits several limitations that need further exploration. First, our evaluation of the generated LCSH relies solely on automatic metrics. This may be skewed; assigning subject headings is an inherently subjective task, and headings that are generated may not be wrong just because they differ from those assigned by human catalogers. Future work will incorporate human evaluation to help assess the relevance and appropriateness of the generated LCSH terms in practical ways. Second, the scope of models used in our experiments is constrained by our available resources. Although models like DeepSeek and the GPT series outperformed Llama-3 in the baseline tests, we were unable to perform experiments applying our optimizations to these models due to constraints in the feasibility of deploying them. Third, other potentially effective methods such as retrieval-augmented generation (RAG), which could use LCSH as an external knowledge base, were not explored in this study. Addressing these limitations will be essential for enhancing the scalability and usability of our framework. In future work, we aim to expand to larger datasets and more advanced LLMs and explore retrieval-augmented generation (RAG) and human feedback to further enhance performance. Moreover, we acknowledge that deploying LLMs in library cataloging and knowledge management systems has many complex ethical implications that this study does not address: copyright, privacy and security, perpetuation of harmful biases, and potential devaluation of catalogers’ labor and expertise, to name a few. Ultimately, the long-term goal of our research is to empower libraries, librarians, catalogers, and other knowledge management workers to harness these automated techniques to build human-in-the-loop systems they can deploy locally that will leverage their expertise and enhance their work, not supplant them. We plan to bring these issues more into focus in future research.

\section{Conclusion}
We’ve proposed a novel hybrid framework that explores how the performance of LLMs in subject analysis for library cataloging systems might be enhanced. Our approach integrates traditional ML models with LLMs. We improved the recall of Llama-3.1-8B from 43\% to 63\%. The implementation of CoT and fine-tuning can enhance the ability of LLMs to generate accurate LCSH terms. Moreover, by integrating the predicted number of terms from a lightweight model as a constraint during generation, we can effectively guide the LLM to generate more concentrated results, resulting in improved precision and F1-score with only a small sacrifice in recall. This offers a flexible trade-off between completeness and accuracy depending on user requirements. Furthermore, the implementation of post-processing to align outputs with terms from LCSH proved to be a highly efficient and low-cost solution for reducing LLM hallucinations. The post-processing step can improve the overall recall without altering the core performance of the underlying language model. Together, these strategies address challenges such as domain adaptation and hallucination. More broadly, the framework and techniques here are generalizable and applicable to a wide range of domain-specific tasks.

\section*{Acknowledgments}
The authors gratefully acknowledge support from the Institute of Museum and Library Services (IMLS). Jinyu Liu, Jason Thomale, and Lingzi Hong received support under Grant \#LG-256666-OLS-24; Xiaoying Song and Lingzi Hong received support under Grant \#LG-256661-OLS-24.

\bibliographystyle{unsrt}  
\bibliography{references}

\end{document}